# IoT-based Remote Control Study of a Robotic Trans-esophageal Ultrasound Probe via LAN and 5G

Shuangyi Wang[1], Xilong Hou[2], Richard Housden[3], Zengguang Hou[1], Davinder Singh[4], Kawal Rhode[3]

[1] Institute of Automation, Chinese Academy of Sciences (CASIA), China
[2] Peng Cheng Laboratory (PCL), China
[3] School of Biomedical Engineering & Imaging Sciences, King's college London, UK
[4] Xtronics Ltd., Gravesend, UK

shuangyi.wang@ia.ac.cn

**Abstract.** A robotic trans-esophageal echocardiography (TEE) probe has been recently developed to address the problems with manual control in the X-ray environment when a conventional probe is used for interventional procedure guidance. However, the robot was exclusively to be used in local areas and the effectiveness of remote control has not been scientifically tested. In this study, we implemented an Internet-of-things (IoT)-based configuration to the TEE robot so the system can set up a local area network (LAN) or be configured to connect to an internet cloud over 5G. To investigate the remote control, backlash hysteresis effects were measured and analysed. A joystick-based device and a button-based gamepad were then employed and compared with the manual control in a target reaching experiment for the two steering axes. The results indicated different hysteresis curves for the left-right and up-down steering axes with the input wheel's deadbands found to be 15° and 8°, respectively. Similar magnitudes of positioning errors at approximately 0.5° and maximum overshoots at around 2.5° were found when manually and robotically controlling the TEE probe. The amount of time to finish the task indicated a better performance using the button-based gamepad over joystick-based device, although both were worse than the manual control. It is concluded that the IoT-based remote control of the TEE probe is feasible and a trained user can accurately manipulate the probe. The main identified problem was the backlash hysteresis in the steering axes, which can result in continuous oscillations and overshoots.

## 1 Introduction

Trans-esophageal echocardiography (TEE) is a widely used imaging modality for diagnosing heart disease and guiding cardiac surgical procedures [1]. A TEE probe usually comprises an electronic interface, a control handle with two concentric wheels, a flexible endoscopic shaft, and a miniaturized ultrasound transducer mounted on the distal tip. During a procedure, the operator is required to manually hold and manipulate the TEE probe on site, which results in several challenges as many cardiac procedures where TEE is utilized are usually accompanied by X-ray fluoroscopy imaging. In this scenario, the operator is required to stand for long periods of time and wear heavy radiation-protection shielding. Evidences have suggested that up to 10% of the radiation from X-ray is still able to pass through the shielding [2]. Moreover, the protection aprons are heavy and can potentially cause orthopedic injuries to echocardiographers, e.g. various potential problems with the spine, hip, knee, and ankle due to the weight of the protective shielding [3]. Other studies also indicate that the sonographers may suffer from an unusually high incidence of musculoskeletal disorders, including eye strain, musculoskeletal pain or injury, repetitive strain injuries and other hazards [4, 5] as the result of holding the probe in one hand while adjusting scanning parameters with the other hand for prolonged periods of time.

Apart from the inconvenience and tedium of the manual control, the need for highly specialized skills is always a barrier for reliable and repeatable acquisition of ultrasound, especially in developing countries. Studies regarding the use of general ultrasound suggest that the lack of training is one of the biggest barriers to reliable image acquisition in low- and middle-income countries [6, 7]. The problem of skills and training for general ultrasound is made worse for TEE by the views of the heart being relatively complicated to understand. Moreover, as cardiac procedures requiring TEE need both a specialized surgeon and an echocardiographer to be available for the entire procedure, the procedures must be scheduled around both of their timetables.

A recently developed robotic system for TEE has made remote control possible [8, 9]. In the previous study, Bluetooth communication has been implemented, which allows the operator to manipulate the probe from a nearby area. However, the previous study has not scientifically demonstrated the validation experiments for doing this with the use of master control devices. Moreover, the configuration limits the use of the robot only in a local area. This meets the need for controlling the robot in the catheter lab environment but may not meet the future need when the robot is required to be used over a long distance. This was challenging in the past but now become feasible with the rapid developments of internet and 5G techniques. In this paper, we report the further developments of the robotic TEE system by implementing an Internet-of-things (IoT) based configuration so that the robot control and data transfer can be achieved over a network. The primary aim of the study is to test the feasibility of controlling the TEE probe remotely using the existing robot via local area internet (LAN) and current emerging 5G. To further understand the effectiveness of remote control, we also aim to investigate the use of gamepad and joystick as the master control device and study the associated problems identified from the user performances.

## 2  Materials and Methods

### 2.1  Robotic TEE Probe

An overview of the robotic system is shown in Fig. 1. The add-on TEE robot holds the probe handle and manipulates four degrees of freedom (DOFs) that are available in manual handling of the probe, including the rotation about and translation along the long axis of the TEE probe and additional manipulators with 2-DOFs to steer the probe head [10]. The overall rotation diameter of the handle control chamber is 110 mm. The dimensions of the whole robot are 500×210×40 mm. The robot is controlled by two microcontrollers with an operating frequency of 16 MHz (Arduino Nano, Adafruit Industries, New York, the United States).

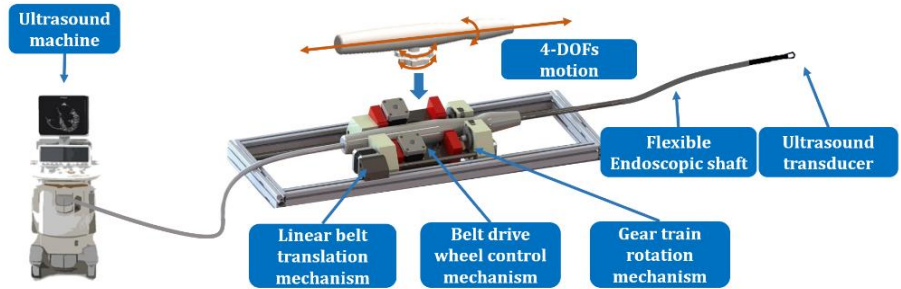

**Fig. 1.** Schematic drawing showing the overview of the proposed robotic TEE probe.

### 2.2 IoT-based Remote Control

The proposed robotic TEE probe working as an IoT device when in clinical use would possibly include the following scenarios (Fig. 2): (1) The robot is placed in the surgical room and the echocardiographer sits in the adjacent control room. A local area network is required for the operator to remotely control the robot and stream the ultrasound images. (2) The echocardiographer can be in centralized hospitals, outpatient clinics, or home environments. In this setup, both the input devices controlled by the echocardiographer and the TEE robot need to be connected into the network and data can be transmitted via the internet cloud.

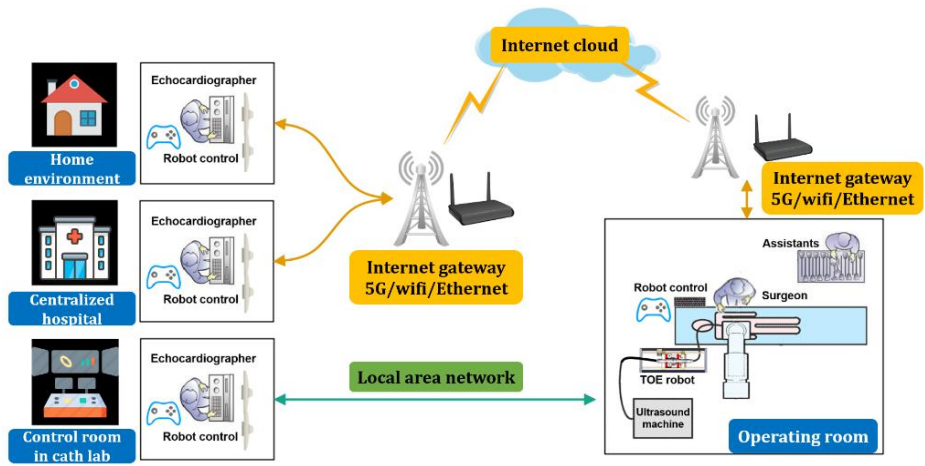

**Fig. 2.** Schematic diagram showing the potential long distance uses of the TEE robot.

To enable the IoT-based control over internet, two TTL-WiFi converter modules (DT-06, Shenzhen Doctors of Intelligence & Technology Co., Ltd., Shenzhen, China) have been built into the robot. The module enables seamless transparent transmission between serial and WiFi and can be easily configured with the built-in HTTP web server. With further developments of the supporting hardware and software, the TEE robot can work in two different modes. In the station (STA) mode, the robot can connect to a WiFi network, e.g. one created by the wireless router or a 5G hotpot. In the access point (AP) mode, the robot can create its own network and have other devices connect to it. In the following experiments, 5G hotpot created by mobile phones are used to create a WiFi network when testing in the STA mode.

In the current study, we aim to test two different input devices. These include a joystick-based device (M5Stick JoyC, M5Stack, Shenzhen, China) which can be directly connected into the network and a button-based gamepad (SN30-G, 8BITDO Tech Co., Ltd., Shenzhen, China) which needs to be connected to a software interface (either on PC or microphone) via Bluetooth first before accessing the network. For both devices, each axis of movement is simply configured either to 'on' or 'off' state. By actuating certain axis on the devices at the master side, the corresponding axis of the TEE robot would be actuated. When an axis on the device is released, the corresponding axis of the TEE robot would be disabled and stopped.

### 2.3 Experimental design

To test the IoT-based remote control of the TEE robot and investigate the associated problems, an experimental setup has been designed (Fig. 3a).

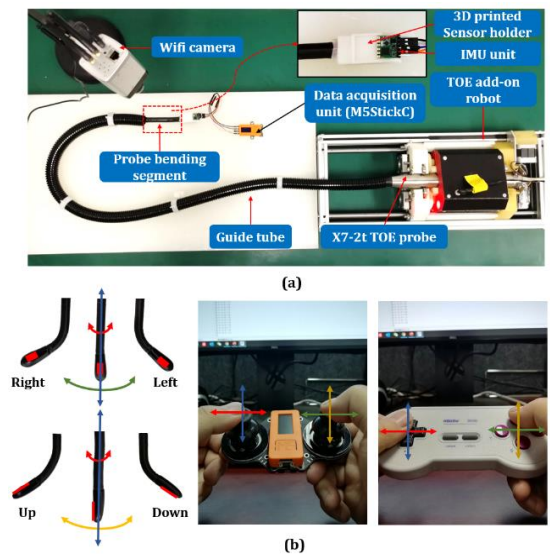

**Fig. 3.** Experimental setup for the remote-control tests: (a) configuration of the robot side setup in a lab; (b) configuration of the control side setup in a separate nearby office.

On the robot side in the lab, the original TEE probe (x7-2t, Philips, The Netherlands) was attached to the add-on robot. A guide tube was fixed to a bench to constrain the shape and provide a path for the flexible endoscopic shaft of the TEE probe. On the tip of the TEE probe where the ultrasound transducer is mounted, an inertial measurement unit (IMU, JY901, Shenzhen Wit Intelligent Co., Ltd., Shenzhen, China) was attached using a 3D printed sensor holder to provide a feedback of the probe tip's pose. A WiFi camera was utilized to provide the real-time video of the probe tip's movements when controlling remotely. On the control side in a separate nearby room, two different input control devices were investigated in the following experiments. The definitions of the motion are shown in Fig. 3b. Real-time video that monitors the probe tip's motion and the IMU readings in the format of three rotational angles were shown on the PC's screen. The following experiments were performed:

1. The first experiment aims to test whether there is a time difference between AP and STA mode when the robot completes a certain task. In this experiment, the AP mode set the robot as the sever while the STA mode utilized an internet cloud (Doit test cloud) as the server. Five different sequential movements of the robot were tested both in the AP and STA modes with each of them includes movements of all four axes of the robot. The amounts of time to finish the task were recorded. In the following experiments, STA mode was then utilized.
2. The second experiment aims to investigate the backlash hysteresis happening to the two steering axes. This is potentially the main influence on the remote control. The wire transmission between the navigation wheels and the bending section induces backlashes into the TEE probe. This can have an impact to the user's control as there are no output responses in the deadbands when reversing the direction. To quantify the size of the deadbands, the probe tip was steered to the leftmost and driven towards the rightmost with 400 motor steps interval. The probe tip was then steered back to the leftmost with the same interval. Same operation was applied to the up-down steering axis starting from the upmost location. IMU angles were recorded at each location and the measurement on each axis was repeated three times. The average angles were then calculated and used to form the hysteresis curves. It should be noted that the flexible endoscopes normally have a complex non-linear behavior [10, 11]. The deadbands of backlashes also depend on the configuration of the flexible endoscopic shaft. The current study measured the hysteresis with the flexible endoscopic shaft configured as the shape shown in Fig. 3.
3. The third experiment aims to explore the users' performances when controlling each steering axis moving towards targets. 10 targets were defined for each axis. These targets are within the realistic ranges that are used in the real TEE acquisition based on our previous analysis [9]. They are specially arranged to include small adjustments, movements passing through hysteresis zone, and directional change outside and within the hysteresis zone. In the experiment, three participants were involved. They are all non-TEE experts but have been very well trained to use the joystick-based device and the button-based gamepad to control the robot. They were asked to control the left-right and up-down steering axes separately to reach the defined 10 targets following the designed sequence. Both devices were tested in this experiment and the participants were given three times for the trial of each axis. The recorded IMU angles and the WiFi camera's video were provided as the feedbacks. The participants were also asked to do manual control trials of the probe to complete the same tasks afterwards.

## 3       Results

By comparing the recorded amount of time to finish each sequence at AP and STA modes, the first experiment has verified there was no obvious time difference between the two working modes. The measured IMU angles in the second experiment were processed and the average hysteresis curves are shown in Fig. 4. The left-right steering axis has a deadband of 1200 motor steps (~15° wheel's rotation angle) that covers the whole steering iteration. The increase and decrease of the IMU angles are mostly linear when steering at a certain direction. The up-down steering axis has a deadband of 640 motor steps (~8° wheel's rotation angle) that only covers a region near the neutral position of the probe tip (1800 to 4200 motor steps; ~-15° to +15° wheel's rotation angle). The increase and decrease of the IMU angles are mostly linear when it is in the hysteresis zone with two different slopes. The angle changes have more complicated non-linear behaviors where the probe tip is steered out of the hysteresis zone.

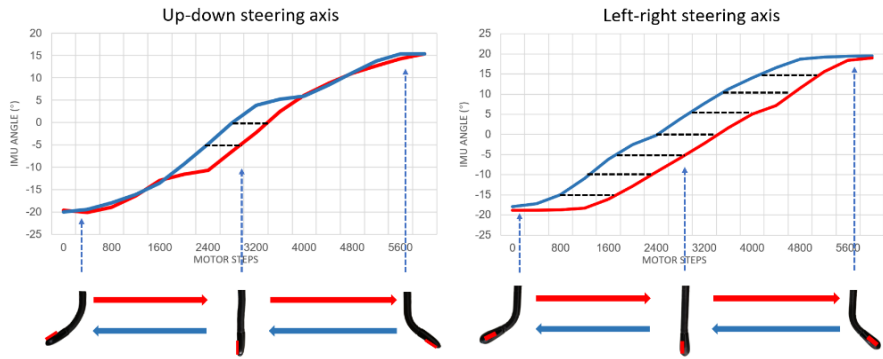

**Fig. 4.** Backlash hysteresis for the two steering axes in the same fixed configuration.

The participants' performances in the third experiment were analysed. An example performance for a single trial is shown in Fig. 5 as an example of the performance curves. In the figure, the blue line shows the recorded IMU angles, the yellow line shows the defined 10 targets, and the red zones highlight the oscillation experienced in the hysteresis deadbands. Although each volunteer's performances are different at each individual trail, several common findings were identified by analysing all three trials of the three participants over the 10 targets (N = 90) on each axis. The results are summarized in Table 1 with the position errors, the maximum overshoots, and the amounts of time to finish each individual task (from one target to the next) calculated.

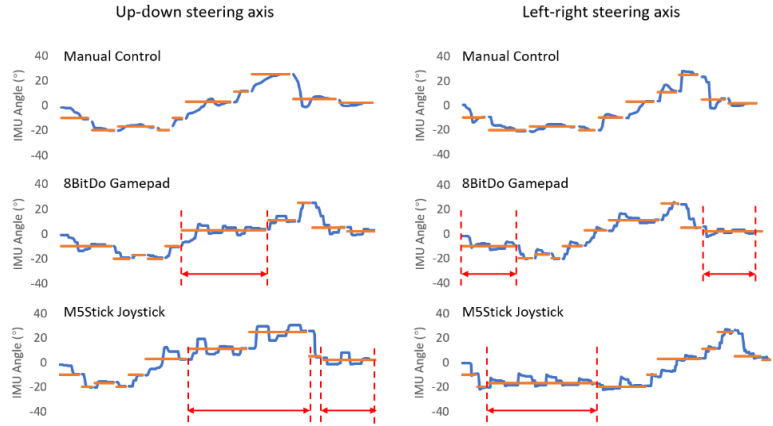

**Fig. 5.** Example participant performance curves for the target reaching experiments under the manual control, the button-based gamepad control, and the joystick-based device control.

As can be seen in Table 1, similar magnitudes of positioning errors at around 0.5° and maximum overshoots at around 2.5° were found when manually and robotically controlling the probe. In terms of the amount of time to finish an individual task, a better performance using the button-based gamepad over joystick-based device has been found with certain level of differences identified ($p$=0.068 for L-R and $p$=0.477 for U-D axis, Welch's t-test), although both were clearly worse than the manual control ($p$=0.002 for L-R and $p$=0.001 for U-D axis, Welch's t-test). The left-right steering axis with a constant hysteresis zone has shown a better result in terms of the data variation compared to the up-down steering axis which has a relatively more complicated hysteresis (Coefficient of Variation = 0.72, 0.87, and 1.29 for U-D axis and 0.37, 0.44, and 0.68 for L-R axis with three different control inputs). By analyzing all the participant's performance curves, oscillation experienced in the hysteresis deadbands are more likely to happen in the up-down steering axes.

**Table 1.** Summary of the participants' performances using different control inputs

| Control Inputs | Error (Mean ± std) | | Maximum overshoot (Mean ± std) | | Time (Median ± IQR) | |
|---|---|---|---|---|---|---|
| | U-D steering | L-R steering | U-D steering | L-R steering | U-D steering | L-R steering |
| Manual | 0.4 ± 0.3° | 0.4 ± 0.2° | 1.1 ± 2.1° | 2.6 ± 2.5° | 6.8 ± 4.9s | 7.0 ± 2.6s |
| Gamepad | 0.5 ± 0.3° | 0.5 ± 0.3° | 2.7 ± 2.6° | 2.3 ± 2.3° | 11.5 ± 10.0s | 9.0 ± 4.0s |
| Joystick | 0.4 ± 0.3° | 0.5 ± 0.3° | 2.6 ± 3.3° | 2.5 ± 3.0° | 12.0 ± 15.5s | 12.5 ± 8.5s |

## 4  Discussion and Conclusion

In this paper, we introduce the development progress with an existing TEE robot by implementing an IoT-based control. It is generally believed that with the developments of internet and 5G techniques, controlling medical robots remotely over a long distance would have the potential to change the current approaches of diagnosis and surgery. For echocardiography, examples have been found in some recent clinical studies for transcatheter procedures guided solely by trans-thoracic echocardiography conducted in China [12, 13]. The clinical team, rather than staying in the centralized hospital in big cities, was routinely deployed to several small clinics in less developed areas to perform the procedure locally. This suggested the future use of echocardiography would need to be remotely over a long distance.

The backlash hysteresis effects and the deadbands are clearly identified as the main problem that would influence the user's performance for the steering axes. Those oscillations identified in the Fig. 6 can be interpreted as the user tried to reach certain targets but had small overshoots. When reversing the direction, the deadbands due to the backlash resulted in a short period of time that the user's inputs are not responded, which then result in further overshoots. Evidences shown in the Fig. 5 and Table 1 also clearly indicate the influence of the hysteresis, especially on the up-down steering axis. It is identified that the manual control of the probe can deal with the hysteresis in a much better way compared to the robotic control. This indicates better control algorithms which can smooth and optimize the speed of the motor when in the deadbands are important in the future developments.

For the joystick-based device and button-based gamepad, the results indicate that well trained users can manipulate the probe to reach a target with a small error and acceptable maximum overshoot. Considering the field of view of the ultrasound image, either in 2D or 3D, it is unlikely that these errors would cause problems to the acquisitions. As only IMU sensor was used in the current study and the combined steering and rotation motions can influence the measured results of IMU, the investigation for the user performances was limited only to the two steering axes. Meanwhile, we believe the rotation axis and translation axis would be easier to control as there are no hystereses. The IMU sensor used in this study is based on the Cortex-M0 core and incorporates novel on-board dynamic processing method with Kalman filter to achieve low-drifts sensing. The rated measurement accuracy is less than 0.1°.

The current study was limited by the number of participants as we hope to only have the participants who are very well trained to use the control devices and the robots, so the results of the remote control are convincing. The study is also limited by not having realistic ultrasound images and TEE experts involved due to the current unavailability. The overall system lag is an important consideration in remote control scenario and a highly accurate method to measure the input and output signals' delay is yet to be developed. These are the problems to solve in our future studies.

To conclude, we implemented an Internet-of-things (IoT)-based configuration to the existing TEE robot so the system can set up a LAN or be configured to access an internet cloud over 5G in this study. The feasibility of the remote control over internet using a joystick-based device and a button-based gamepad has been verified. It is believed that a trained user can accurately manipulate the probe with tolerable overshoots. The main identified problem was the backlash hysteresis in the steering axes which can result in continuous oscillations and overshoots. The future work should focus on implementing advanced control algorithm to deal with the hysteresis and test the remote control with realistic ultrasound images and TEE experts involved.